\documentclass[a4paper,twoside]{article}

\usepackage{calc}
\usepackage{amssymb}
\usepackage{amstext}
\usepackage{amsmath}
\usepackage{amsthm}
\usepackage{multicol}
\usepackage{pslatex}
\usepackage{apalike}
\usepackage{SCITEPRESS}
\usepackage[small]{caption}

\usepackage{graphicx}
\usepackage{caption}
\usepackage{wrapfig}
\usepackage{float}
\restylefloat{table}
\usepackage{floatflt}

\newcommand{\bmx}[0]{\begin{bmatrix}}
\newcommand{\emx}[0]{\end{bmatrix}}

\newcommand{\qlay}[1]{\left[#1\right]}
\newcommand{\vect}[1]{\mathbf{#1}}
\newcommand{\vects}[1]{\boldsymbol{#1}}
\newcommand{\matr}[1]{\mathbf{#1}}

\newcommand{\vh}[0]{\vect{h}}

\newcommand{\vx}[0]{\vect{x}}

\newcommand{\vs}[0]{\vect{s}}
\newcommand{\vf}[0]{\vect{f}}
\newcommand{\vy}[0]{\vect{y}}

\newcommand{\mW}[0]{\matr{W}}

\newcommand{\mU}[0]{\matr{U}}

\newcommand{\TT}[0]{{\vects{\theta}}}

\newcommand{\LL}[0]{\mathcal{L}}

\DeclareMathOperator*{\argmax}{arg\,max}

\usepackage{url}
\urldef{\mails}\path|{firstname.lastname@aalto.fi}|

\begin{document}

\title{Classifying and Visualizing Motion Capture Sequences
\\using Deep Neural Networks }

\author{\authorname{Kyunghyun Cho and Xi
Chen\sup{1}}
\affiliation{Department of Information and Computer
Science, Aalto University School of Science, Espoo, Finland}
\email{\{kyunghyun.cho,xi.chen\}@aalto.fi}
}


\abstract{
The gesture recognition using motion capture data and depth
sensors has recently drawn more attention in vision
recognition.  Currently most systems only classify dataset
with a couple of dozens different actions. Moreover, feature
extraction from the data is often computational complex.  In
this paper, we propose a novel system to recognize the
actions from skeleton data with simple, but effective,
features using deep neural networks.  Features are extracted
for each frame based on the relative positions of joints
(PO), temporal differences (TD), and normalized trajectories
of motion (NT).  Given these features a hybrid multi-layer
perceptron is trained, which simultaneously classifies and
reconstructs input data.  We use deep autoencoder to
visualize learnt features. The experiments show that
deep neural networks can capture more discriminative
information than, for instance, principal component analysis
can.  We test our system on a public database with 65
classes and more than 2,000 motion sequences. We obtain an
accuracy above 95\% which is, to our knowledge, the state of
the art result for such a large dataset. }

\onecolumn \maketitle \normalsize \vfill

\section{\uppercase{Introduction}}
\label{sec:introduction}

\footnotetext[1]{Both authors contributed equally.}
\addtocounter{footnote}{1}

\noindent Gesture recognition has been a hot and challenging research
topic for several decades. There are two main kinds of
source data: video and motion capture data (Mocap).  Mocap
records the human actions based on the human skeleton
information.  Its classification is very important in
computer animation, sports science, human--computer
interaction (HCI) and filmmaking.

Recently the low cost and high
mobility of RGB-D sensors, such as Kinect, have become widely adopted
by the game industry as well as in HCI. Especially in
computer vision, the gesture recognition using data from the
RGB-D sensors is gaining more and more attention.  However,
the computational difficulty in directly processing 3--D
cloud data from depth information often leads to utilizing the
human skeleton extracted from the depth information 
\cite{Shotton_real-timehuman} instead.

However, conventional recognition systems are mostly applied on a small 
dataset with a couple of dozens different actions, which is often
the limitation imposed by the design of a system. Conventional designs 
may be classified into two categories: 
a whole motion is represented by one feature matrix or vector \cite{raptis2011real}, and 
classified by a classifier as a whole \cite{muller2006motion}; or a library of  
key features \cite{wang2012mining} is learned from the whole dataset, 
and then each motion is represented as a bag or histogram of words \cite{raptis2008flexible} 
or a path in a graph \cite{barnachon2013real}. 

In the first type of system, principal component analysis (PCA) is 
often used to form equal-size
feature matrices or vectors from variable-length motion sequences 
\cite{Zhao201345,vieiradistance}. However, due to a large number of 
inter- and intra-class variations
among motions, a single feature matrix
or vector is likely not enough to capture important discriminative 
information. This makes these systems inadequate for a large dataset.

The second type of system decomposes a motion with
a manual setup sliding window or key features \cite{raptis2008flexible} 
and builds a codebook by clustering 
\cite{CRF2013}. These approaches also suffer from a large number of action 
classes due to a potentially excessive size of codebook, in the case of 
using a classifier such as support vector machines \cite{MHAD2013} 
as well as an overly complicated structure, if one tries to build a motion graph.

In this paper, we recognize actions from  skeleton data with 
two major contributions: $(1)$ we propose to build the 
recognition system based on joint distribution model of the 
per-frame feature set containing information of the relative positions of joints, 
their temporal difference and the normalized trajectory of the motion;
$(2)$ we propose a novel variant of a multi-layer perceptron, 
called a hybrid MLP, that simultaneously classifies and reconstructs 
the features, which outperforms
a regular MLP, extreme learning machines (ELM) as well as SVM; meanwhile  
a deep autoencoder is trained to visualize the features in 
2-dimensional space, comparing with PCA using the two leading 
principal components, we clearly see that autoencoder can 
extract more distinctive information from features. 
We test our system on a publicly available database 
containing 65 action classes and
more than 2,000 motions  with above 95\% accuracy.
\begin{figure*}[t]
\centering 
 \begin{minipage}{0.44\textwidth}
     \centering
     \includegraphics[width=1\columnwidth]{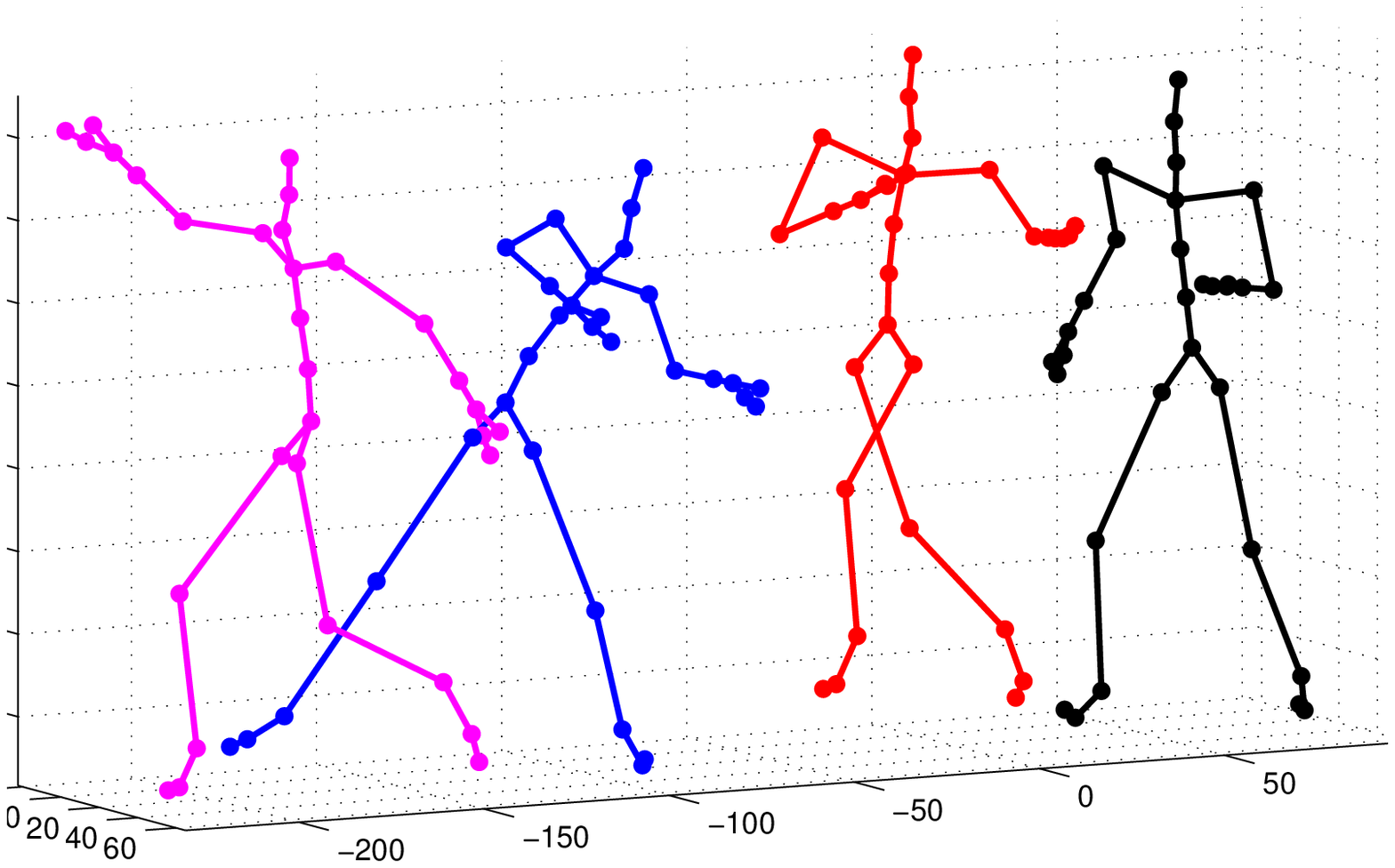} 
     \\
     \scriptsize
     (a) Original
 \end{minipage}
 \begin{minipage}{0.44\textwidth}
     \centering
     \includegraphics[width=0.9\columnwidth]{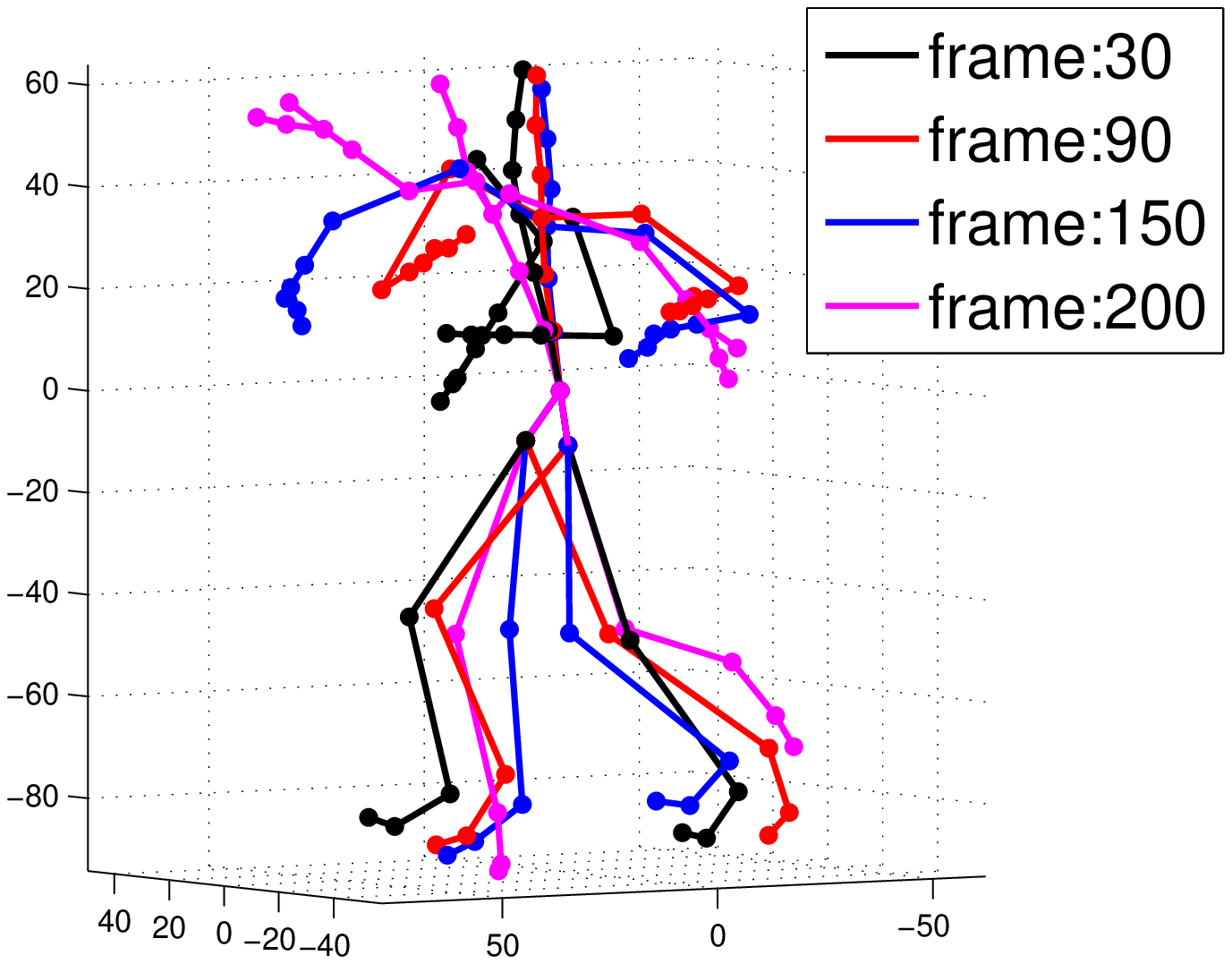}
     \\
     \scriptsize
     (b) Root
 \end{minipage}
 \caption{A \textit{throwFar} action in (a) original and (b) root
coordinates.}
\label{fig:throwfar}
\end{figure*}

\section{\uppercase{Feature Extraction}}
\label{sec:features}
\noindent In this section we describe how to extract the proposed features from
each frame.
Fig.~\ref{fig:throwfar} (a) shows some frames from a motion
sequence \textit{throwfar}. Since the original coordinates
are dependent on a performer and the space to which the
performer belongs, those coordinates are not directly
comparable between different performers even if they all
perform the same action. Hence, we normalize the orientation 
such that each and every skeleton has its root at the origin
$(0,0,0)$ with the same orientation matrix of identity. 
For example, Fig.~\ref{fig:throwfar} (b) shows
the orientation-normalized versions of the skeleton in
Fig.~\ref{fig:throwfar} (a). We further normalize the
length of all connected joints to be $1$ to make them
independent of a performer. The concatenation of 3D 
coordinates of joints forms the feature (PO), which describes 
relative relationships among the joints. 

\begin{figure*}[t]
\centering 
 \begin{minipage}{0.44\textwidth}
     \centering
     \includegraphics[width=0.9\columnwidth]{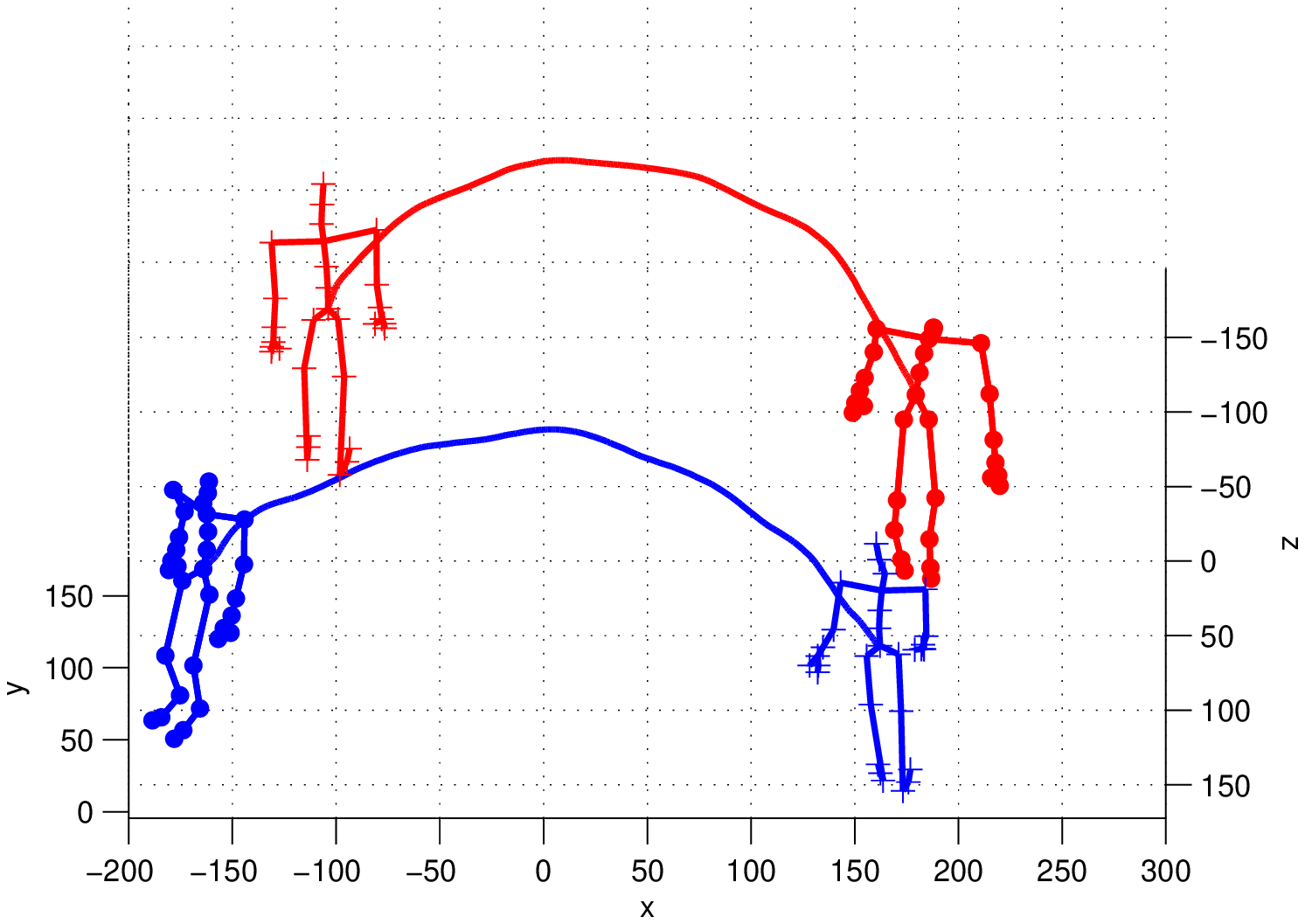}
     \\
     \scriptsize
     (a) Original
 \end{minipage}
 \begin{minipage}{0.44\textwidth}
     \centering
     \includegraphics[width=0.9\columnwidth]{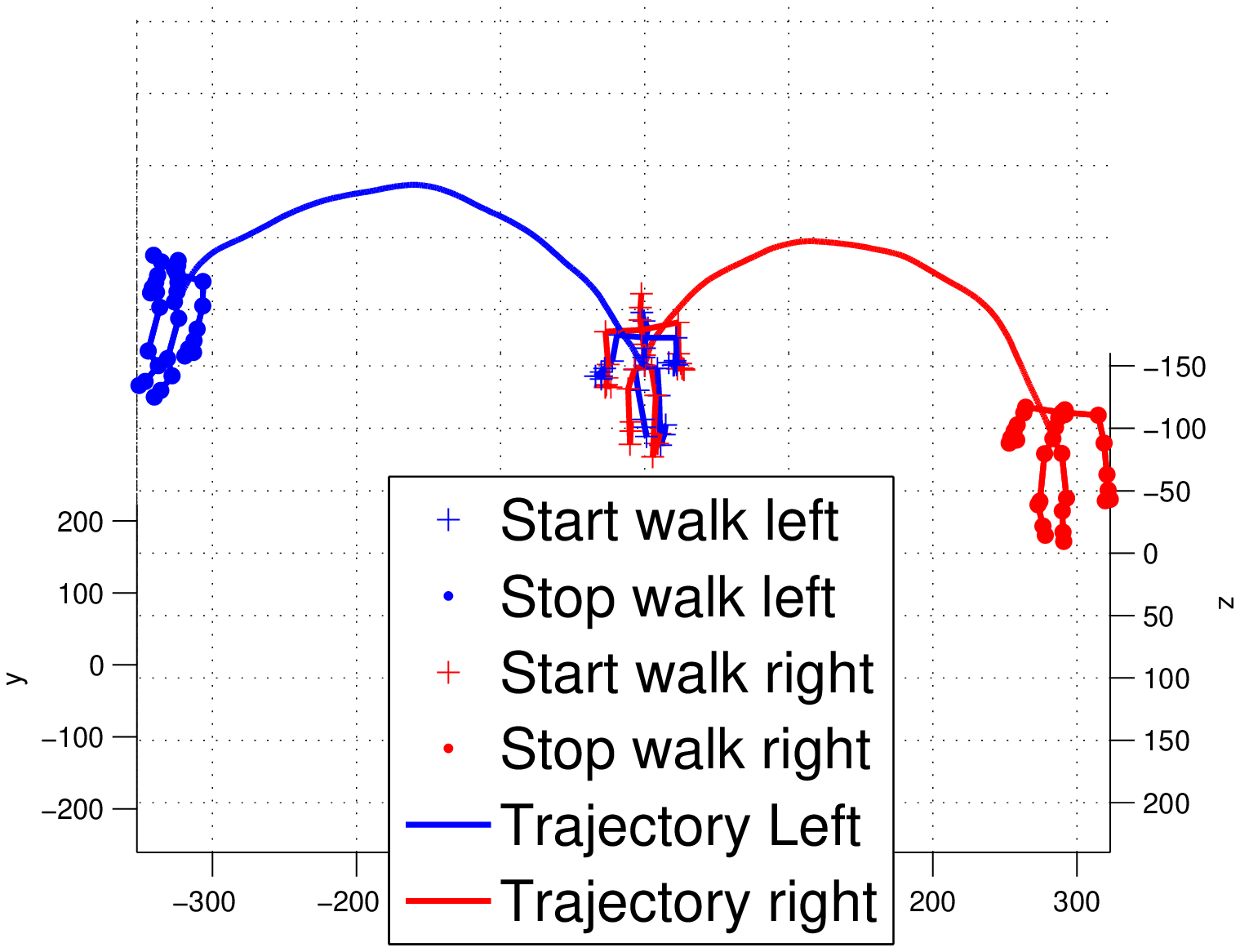}
     \\
     \scriptsize
     (b) Transformed
 \end{minipage} 
 \caption{Trajectories of two different walks in (a) original
and (b) transformed coordinates.}
\label{fig:traj}
\end{figure*}

Some actions are similar
to each other in a frame-level. For instance, the actions of
standing up and sitting down are just reverse in time but
with almost identical frames, which results in almost
identical PO features for corresponding frames in those
actions. Hence, we compute the temporal differences (TD) between
pairs of PO feature  by
\begin{equation} \label{eq:tddv}
    \mathbf{f}_{\text{TD}}^i=\left\{ 
  \begin{array}{l l}
      \mathbf{f}_{\text{PO}}^i & \quad \text{$1 \leq i<m$ } \\
      \dfrac{\left(\mathbf{f}_{\text{PO}}^i -
      \mathbf{f}_{\text{PO}}^{i-m+1}\right) }
      { ||\mathbf{f}_{\text{PO}}^i -
      \mathbf{f}_{\text{PO}}^{i-m+1}||} & \quad \text{$m \leq i \leq N$ },
  \end{array} \right.
\end{equation}
where 
$\mathbf{f}^i_{\text{PO}}$ and $m$ are the PO feature vector at the
$i$-th frame and the temporal offset ($1 < m < N$), respectively.
TD feature preserves the temporal relationship of the same joint.
Normalized trajectory (NT) extracts the absolute trajectory of the 
motion. Fig.\ref{fig:traj}(a) shows two motions: 
\textit{walk in a left circle} and \textit{walk in a right circle}. 
However, in this figure the trajectories of left circle and right 
circle are not distinguishable. 
In order to incorporate the trajectory information, 
we set the same orientation and starting position for the root in the 
first frame and use a relative position of the root of all the rest 
frames in the motion sequences from the initial frame, normalized into
$\left[-1, 1\right]$ in each dimension.
See Fig.~\ref{fig:traj} (b) for the effect of this
transformation. 
The final feature for each frame is a concatenation of three features. 
The dimension of the feature is $3\times n \times 2 +3$, where 
$n$ is the number of joints in use in PO. 

For skeleton extracted from RGB-D sensor, often the rotation matrix and 
translation vector related with the joints are not available. 
In this case any skeleton can be selected as a stardard template frame, 
the rotation matrix between the other skeletons and the template 
can be calculated as in \cite{MocapELM}. In the similar way the features 
from skeleton data with only 3D joint coordinates can be extracted. 

\section{\uppercase{Deep Neural Networks: Multi-layer Perceptrons}}
\noindent A multi-layer perceptron (MLP) is a type of deep neural
networks that is able to perform classification (see, e.g.,
\cite{Haykin2009}). An MLP can approximate any smooth,
nonlinear mapping from a high dimensional sample to a class
through multiple layers of hidden neurons.

The output or prediction of an MLP having $L$ hidden layers
and $q$ output neurons given a sample $\vx$ is typically
computed by
\begin{align}
    \label{eq:mlp_output}
    u(&\vx \mid \TT) = 
    \\
    &\sigma \left(\mU^\top \phi \left(
    \mW_{\qlay{L}}^\top \phi \left( \mW_{\qlay{L-1}}^\top
    \cdots \phi \left( \mW_{\qlay{1}}^\top \vx \right)
    \cdots \right) \right)\right),
    \nonumber
\end{align}
where $\sigma$ and $\phi$ are component-wise nonlinear
functions, and $\TT=\left\{ \mU, \mW_{\qlay{1}}, \dots
,\mW_{\qlay{L}} \right\}$ is a set of parameters. We have
omitted a bias without loss of generality.
A logistic sigmoid function is usually used for the last
nonlinear function $\sigma$. Each output neuron 
corresponds to a single class.

Given a training set $\left\{ \left( \vx^{(n)}, \vy^{(n)}
\right) \right\}_{n=1}^N$, an MLP is trained to approximate
the posterior probability $p(y_j=1 \mid \vx)$ of each output
class $y_j$ given a sample $\vx$ by maximizing the
log-likelihood
\begin{align}
    \LL_{\text{sup}}(\TT) =& \sum_{n=1}^N \sum_{j=1}^q
    \left(y^{(n)}_j
    \log u_j\left(\vx^{(n)}\right)
    \right.
    \nonumber
    \\
    \label{eq:mlp_ll_sup}
    &+\left.
    \left(1-y_j^{(n)}\right)
    \log\left( 1 - 
    u_j\left(\vx^{(n)}\right)\right)\right), 
\end{align}
where a subscript $j$ indicates the $j$-th component. We
omitted $\TT$ to make the above equation uncluttered.
Training can be efficiently done by backpropagation
\cite{Rumelhart1986}.

\subsection{Hybrid Multi-Layer Perceptron}
\noindent It has been noticed by many that it is not trivial to train
deep neural networks to have a good generalization
performance (see, e.g., \cite{Bengio2007a} and references
therein), especially when there are many hidden layers
between input and output layers. One of promising hypotheses
explaining this difficulty is that backpropagation applied
on a deep MLP tends to utilize only a few top layers
\cite{Bengio2007nips}. A method of layer-wise pretraining
has been proposed 
to overcome this problem by initializing the weights in
lower layers with unsupervised learning \cite{Hinton2006}.

Here, we propose another strategy that forces
backpropagation algorithm to utilize lower layers. The
strategy involves training an MLP to classify and
reconstruct simultaneously by training a deep autoencoder
with the same set of parameters, except for the weights
between the penultimate and output layers to reconstruct an
input sample as well as possible. 

A deep autoencoder is a symmetric feedforward neural network
consisting of an encoder
\[
    \vh = f(\vx) = f_{
    \qlay{L-1}} \circ f_{\qlay{L-2}} \circ \cdots \circ
    f_{\qlay{1}}(\vx)
\]
and a decoder
\[
    \tilde{\vx} = g(\vx) = g_{\qlay{2}} \circ \cdots \circ
    g_{\qlay{L-1}} (\vh),
\]
where 
\begin{align*}
    f_{\qlay{l}}(\vs_{\qlay{l-1}}) &= \phi
    \left(\mW_{\qlay{l}}^\top \vs_{\qlay{l-1}}\right),
    \\
    g_{\qlay{l}}(\vs_{\qlay{l+1}}) &= \varphi
    \left(\mW_{\qlay{l}} \vs_{\qlay{l+1}} \right).
\end{align*}
$\phi$ and $\varphi$ are component-wise nonlinear functions.

The parameters of a deep autoencoder is estimated by
maximizing the negative squared difference which is defined
to be
\begin{align}
    \label{eq:mlp_ll_unsup}
    \LL_{\text{unsup}}(\TT) =  -\frac{1}{2} \sum_{n=1}^N \left\| \vx^{(n)} -
    \tilde{\vx}^{(n)} \right\|_2^2.
\end{align}

Our proposed strategy combines these two networks while
sharing a single set of parameters $\TT$ by optimizing a
weighted sum of Eq.~\eqref{eq:mlp_ll_sup} and
Eq.~\eqref{eq:mlp_ll_unsup}:
\begin{align}
    \label{eq:mlp_ll_simul}
    \LL(\TT) = (1 - \lambda) \LL_{\text{sup}}(\TT) + \lambda
    \LL_{\text{unsup}}(\TT),
\end{align}
where $\lambda \in \left[ 0, 1\right]$ is a hyperparameter.
When $\lambda$ is $0$, the trained model will be purely an
MLP, while it will be an autoencoder if $\lambda = 1$. 
We
call an MLP that was trained with this strategy with
non-zero $\lambda$ a \textit{hybrid} MLP\footnote{ A similar
approach was proposed in \cite{Larochelle2008} in the case
of restricted Boltzmann machines.}.

There are two advantages in the proposed strategy.  First,
the weights in lower layers naturally have to be utilized,
since those weights must be adapted to reconstruct an input
sample well. 
This may further help achieving a better
classification accuracy similarly to the way unsupervised
layer-wise pretraining which also optimized the
reconstruction error
, in the case of using autoencoders,
helps obtaining a better classification accuracy on novel
samples.
Secondly, in this framework, it is trivial to use vast
amount of unlabeled samples in addition to labeled samples.
If stochastic backpropagation is used, one can compute the
gradients of $\LL$ by combining the gradients of
$\LL_{\text{sup}}$ and $\LL_{\text{unsup}}$ separately using
separate sets of labeled and unlabeled samples.
\begin{figure*}[t]
    \centering
    \begin{minipage}{0.44\textwidth}
        \centering
        \includegraphics[width=\columnwidth,clip=true,trim=80 50 70 30]{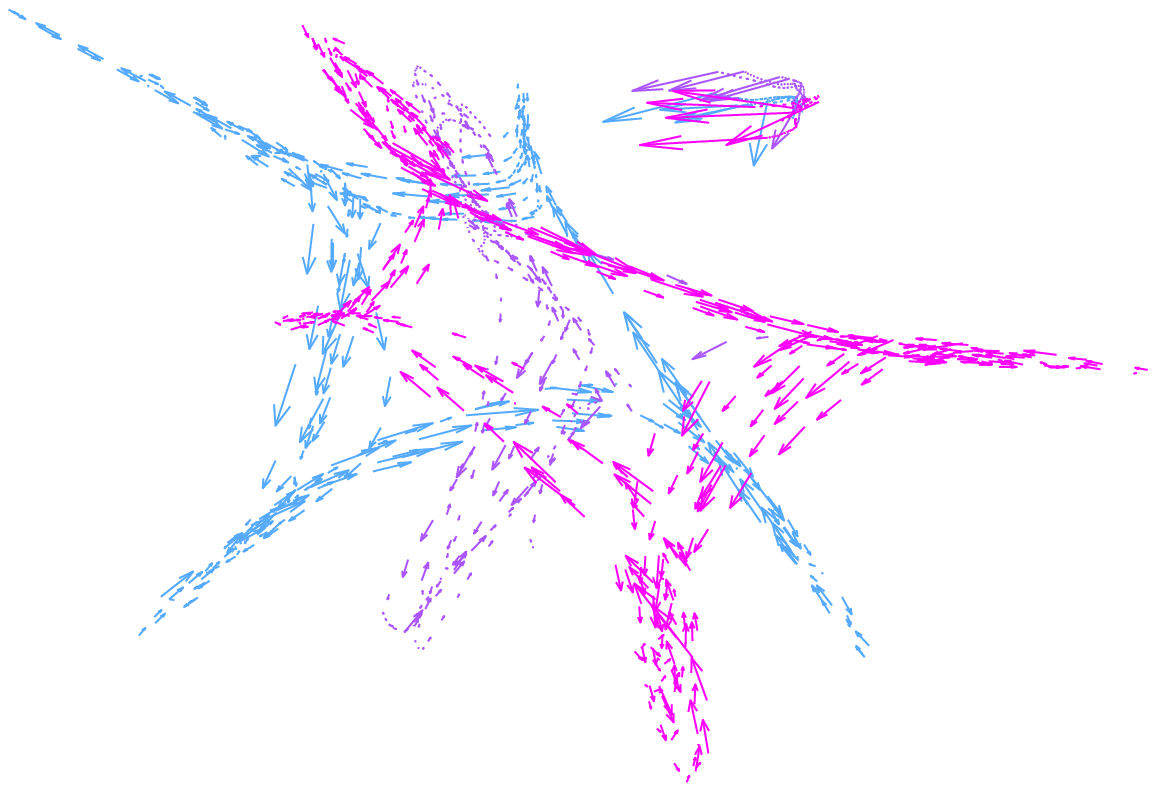}
        \\
        \small
        (a) DNN (PO+TD)
    \end{minipage}
    \begin{minipage}{0.44\textwidth}
        \centering
        \includegraphics[width=\columnwidth,clip=true,trim=80 50 70 30]{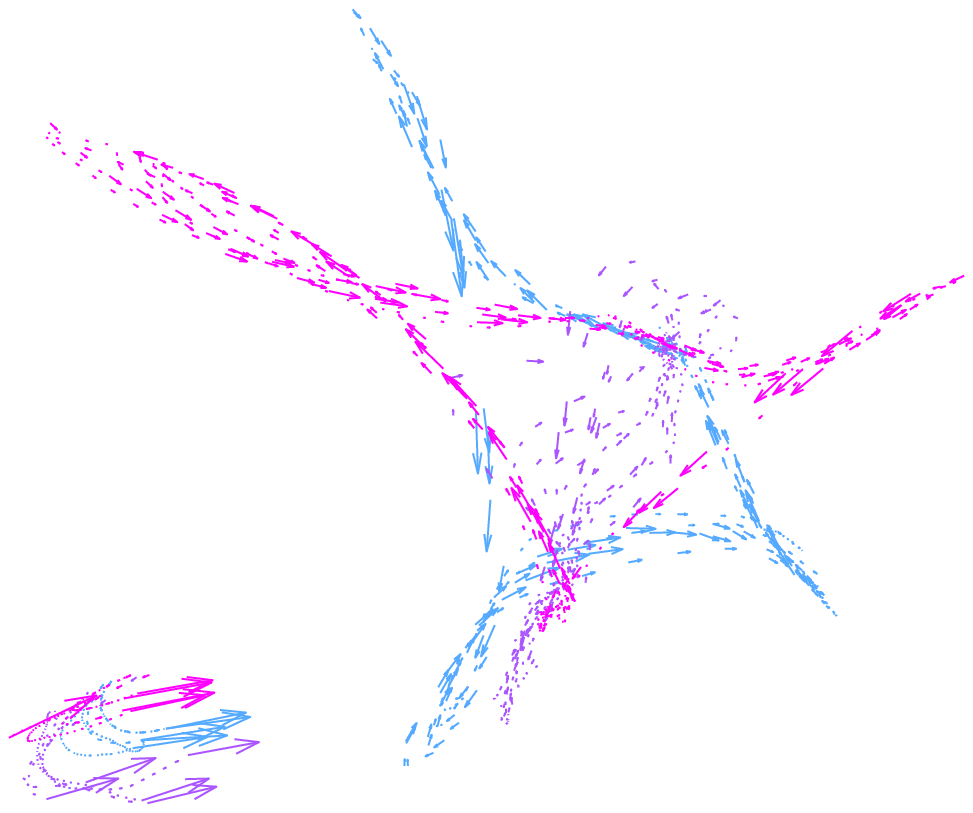}
        \\
        \small
        (b) DNN (PO+TD+NT)
    \end{minipage}
    \\
    \begin{minipage}{0.44\textwidth}
        \centering
        \includegraphics[width=\columnwidth,clip=true,trim=80 50 70 30]{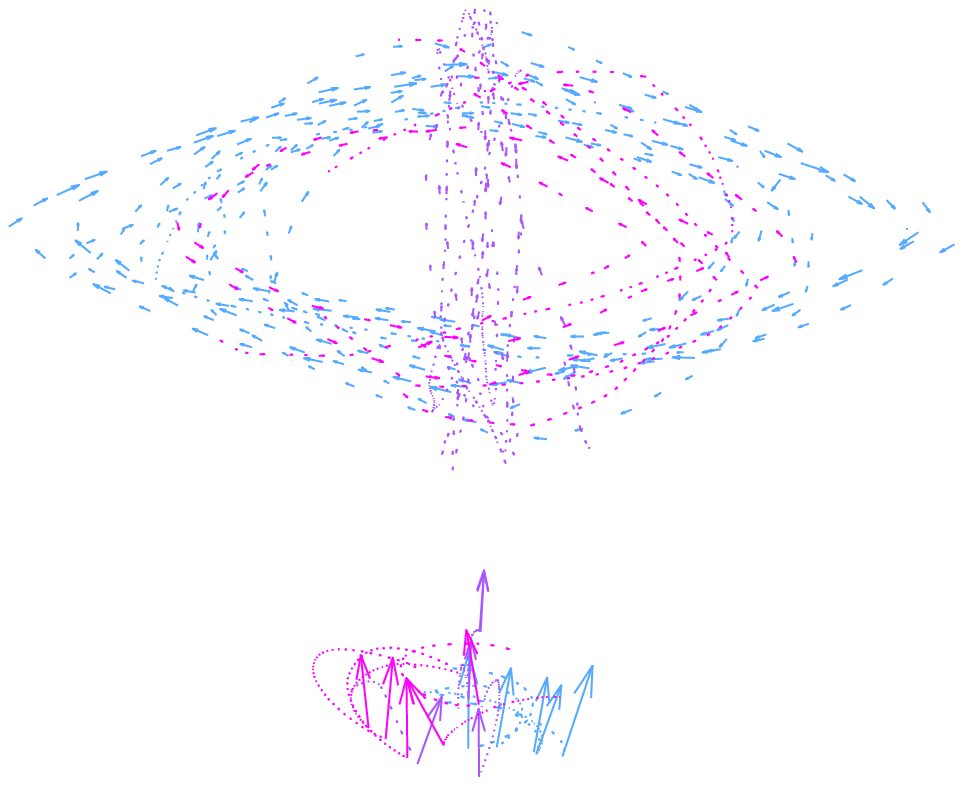}
        \\
        \small
        (c) PCA (PO+TD)
    \end{minipage}
    \begin{minipage}{0.44\textwidth}
        \centering
        \includegraphics[width=\columnwidth,clip=true,trim=80 50 70 30]{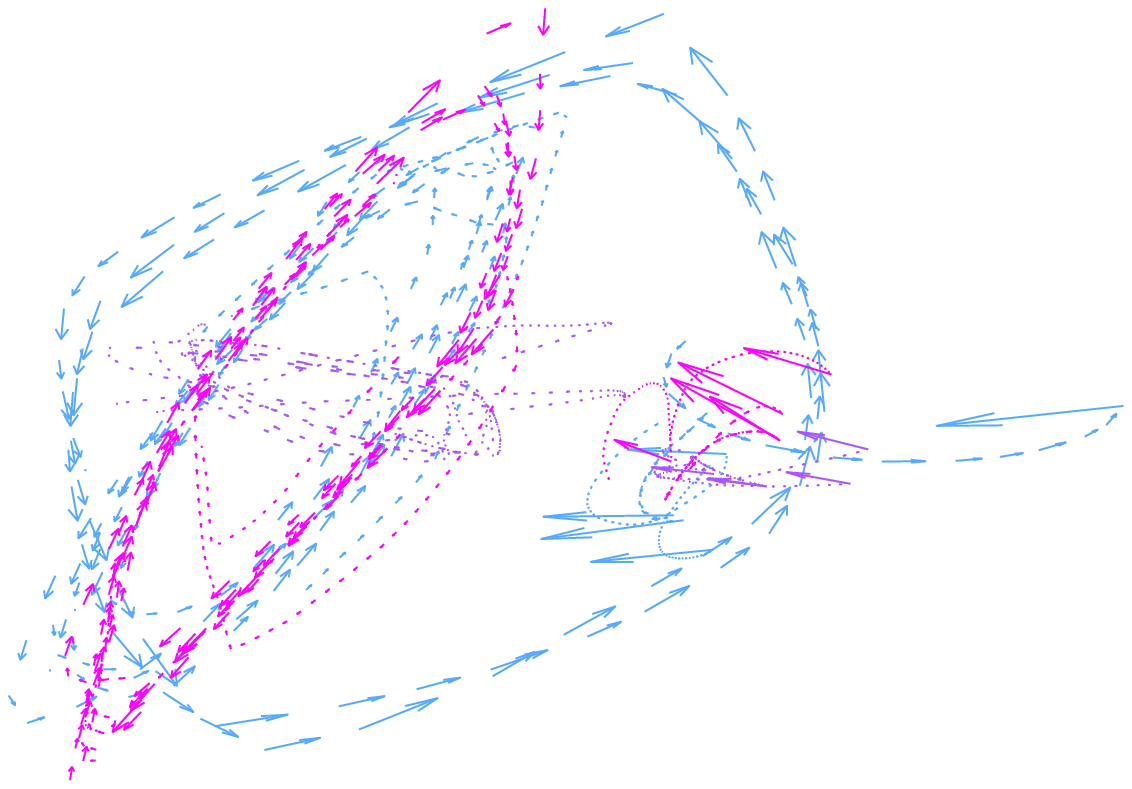}
        \\
        \small
        (d) PCA (PO+TD+NT)
    \end{minipage}
    \caption{Visualization of actions
    \textit{rotateArmsRBackward} (blue),
    \textit{rotateArmsBothBackward} (purple) and
    \textit{rotateArmsLBackward} (red). Each arrow denotes
    the direction and magnitude of change in the latent
    space. 
    Five randomly selected sequences per action are
    shown.
    }
    \label{fig:vis_rotate}
\end{figure*}

\section{\uppercase{Classifying an Action Sequence}}
\noindent An action sequence is composed of an certain amount of frames. 
We use a multi-layer perceptron to model a posterior distribution 
of classes given each frame.
Let us define $s_c \in \left\{ 0, 1\right\}$ be a binary
indicator variable. If $s_c$ is one, the sequence belongs to
the action $c$, and otherwise, belongs to another action.
Since each sequence consists of $N \geq 1$ frames, let us
further define $f_{i,c} \in \left\{ 0, 1\right\}$ as a
binary variable indicating whether the $i$-th frame belongs
to the action $c$. 

When a given sequence $\vs=(\vf_1, \vf_2, \dots , \vf_N)$ is
of an action $c$, every frame $\vf_i$ in the sequence is
also of an action $c$. In other words, if $s_c = 1$,
$f_{i,c} = 1$ for all $i$. So, we may check the joint
probability of all frames in the sequence to determine the
action of the sequence:
\begin{align}
    \label{eq:seq_prob}
    p(&s_c = 1 \mid \vs) = 
    \\
    &p(f_{1,c}=1, f_{2,c}=1, \dots,
    f_{N,c} = 1 \mid \vf_1, \vf_2, \dots, \vf_N).
    \nonumber
\end{align}

In this paper, we assume temporal independence among the
frames in a single sequence, which means that the 
class of each frame depends \textit{only} on the features of
the frame. Then, Eq.~\eqref{eq:seq_prob} can be simplified
into
\begin{align}
    \label{eq:seq_prob_simple}
    p(s_c = 1 \mid \vs) = \prod_{i=1}^N p(f_{i,c}=1\mid
    \vf_i).
\end{align}


With this assumption, the problem of gesture recognition is
reduced to first train a classifier to perform frame-level
classification and then to combine the output of the
classifier according to Eq.~\eqref{eq:seq_prob_simple}. A
multi-layer perceptron which approximates the posterior
probability distribution over classes by
Eq.~\eqref{eq:mlp_output} is naturally suited to this
approach.

\begin{figure*}[t]
    \centering
    \begin{minipage}{0.44\textwidth}
        \centering
        \includegraphics[width=\columnwidth,clip=true,trim=80 50 70 30]{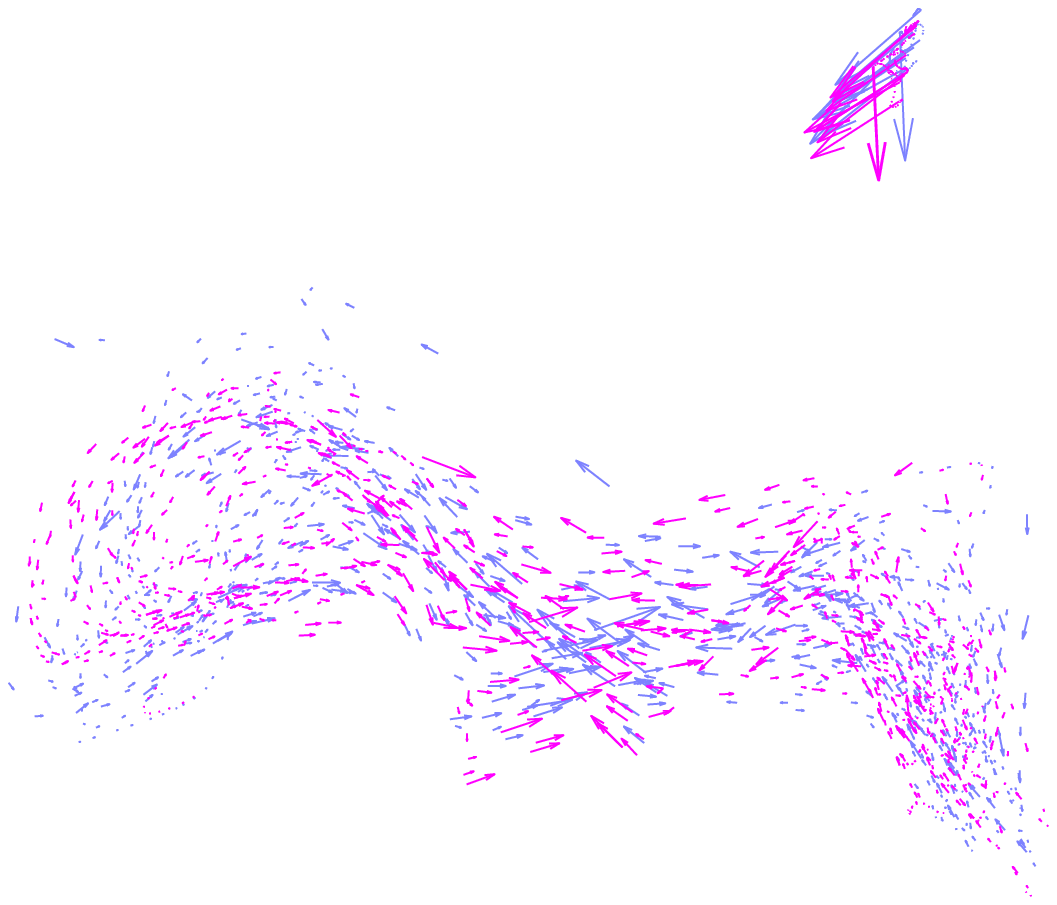}
        \\
        \small
        (a) DNN (PO+TD)
    \end{minipage}
    \begin{minipage}{0.44\textwidth}
        \centering
        \includegraphics[width=\columnwidth,clip=true,trim=60 50 70 30]{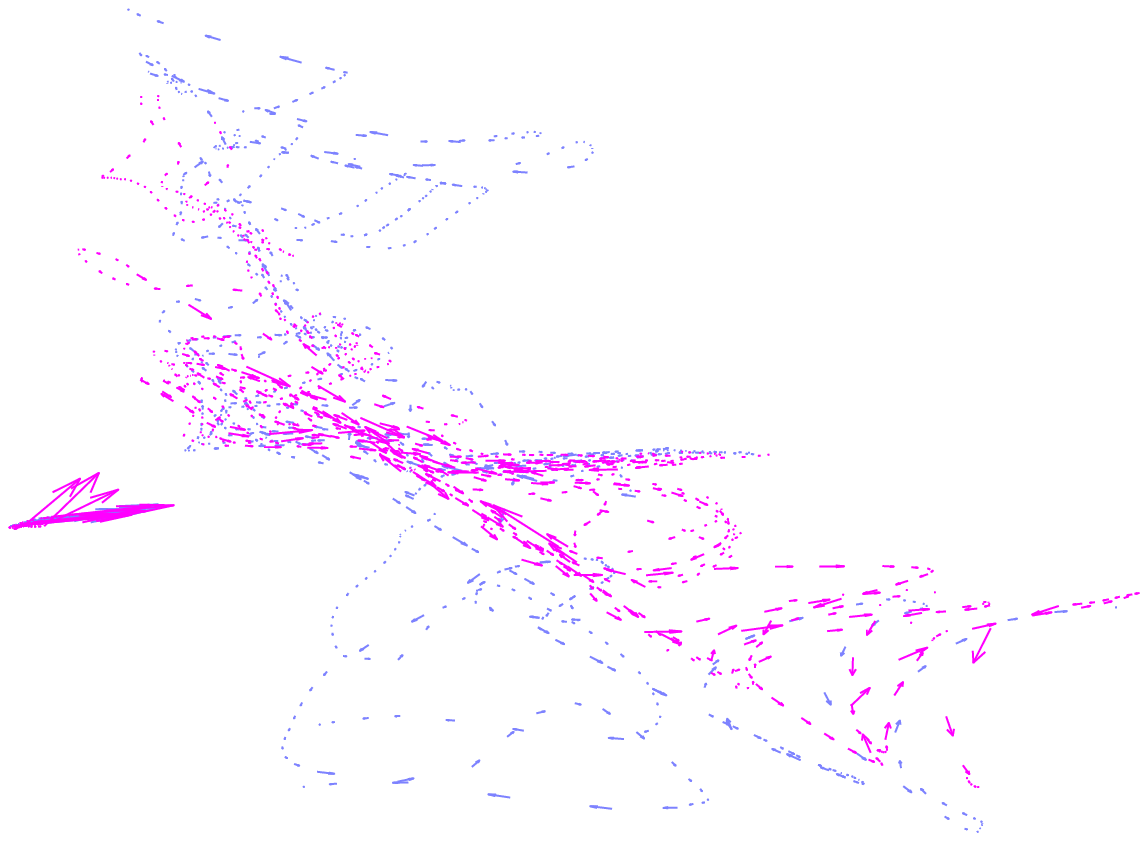}
        \\
        \small
        (b) DNN (PO+TD+NT)
    \end{minipage}
    \\
    \begin{minipage}{0.44\textwidth}
        \centering
        \includegraphics[width=\columnwidth,clip=true,trim=80 50 70 30]{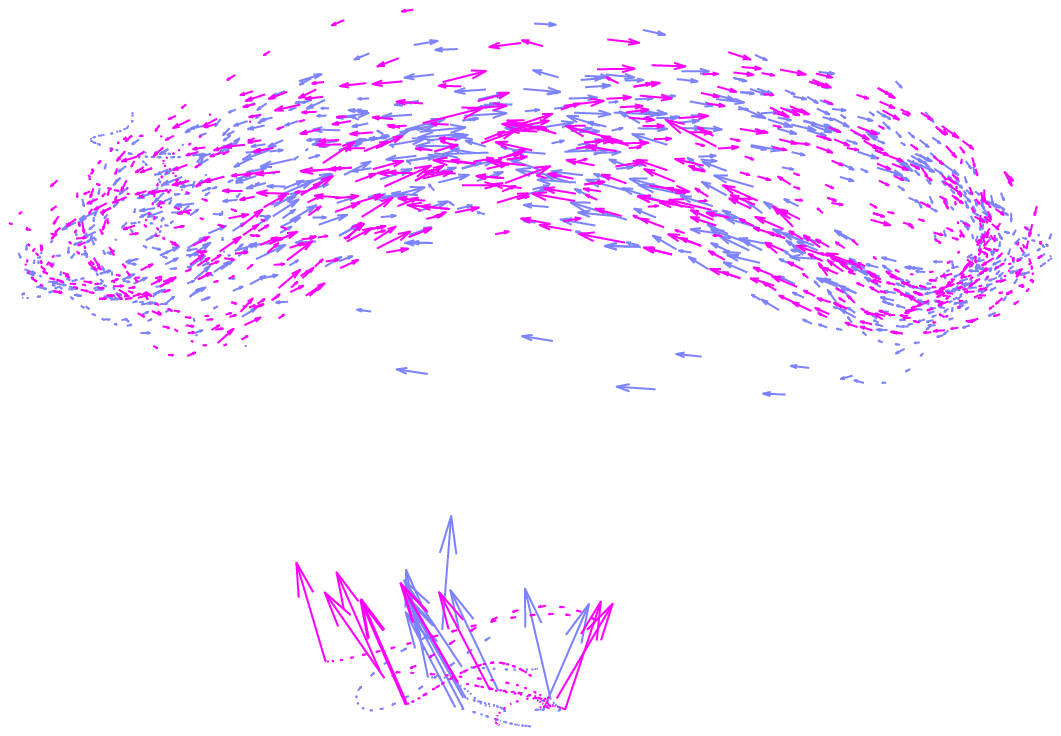}
        \\
        \small
        (c) PCA (PO+TD)
    \end{minipage}
    \begin{minipage}{0.44\textwidth}
        \centering
        \includegraphics[width=\columnwidth,clip=true,trim=80 50 70 30]{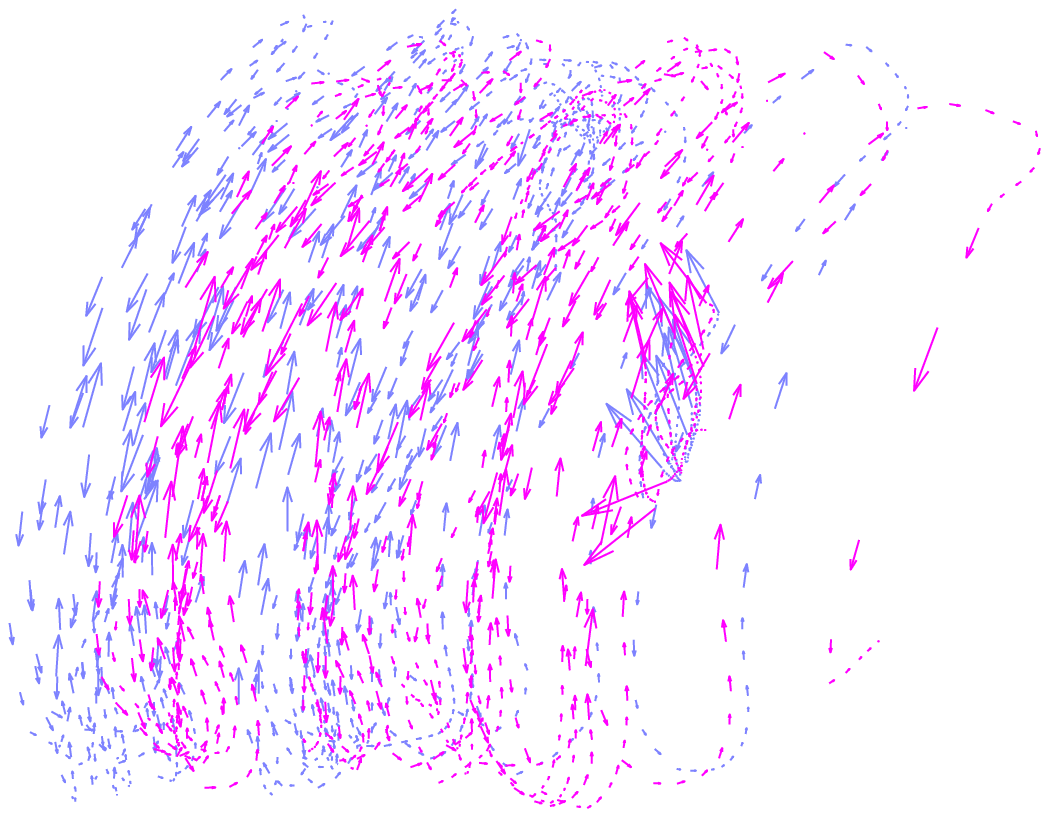}
        \\
        \small
        (d) PCA (PO+TD+NT)
    \end{minipage}
    \caption{Visualization of actions
    \textit{jogLeftCircle} (blue) and
    \textit{jogRightCircle} (purple).
    Each arrow denotes
    the direction and magnitude of change in the latent
    space. 
    Ten randomly chosen sequences per action were
    visualized.
    }
    \label{fig:vis_jog}
\end{figure*}

\section{\uppercase{Experiments}}

\noindent In the experiment we tried to evaluate the performance of
our proposed recognition system through a public dataset.
We assessed the performance of deep neural networks
including regular and hybrid MLP by comparing them against
extreme learning machines (ELM) and support vector machines
(SVM).  The effectiveness of the feature set was evaluated
by the classification accuracy and the visualization in 2D
space by deep autoencoders. 

\subsection{Dataset}

\noindent The Motion Capture Database HDM05~\cite{cg-2007-2} is a well
organized large MOCAP dataset. It provides a set of short
cut MOCAP clips, and each clip contains one integral motion.
In the original dataset, there are 130 gesture groups.
However, there are some gestures that essentially belong to
a single class.  For instance, \textit{walk 2 steps} and
\textit{walk 4 steps} belong to a single action
\textit{walk}. Hence, we combined some of the classes based
on the following rules:
\begin{enumerate}
 \item Motions repeating with different times are combined into one action. 
 \item Motions with the only difference of the starting limb are combined 
 into one action. 
\end{enumerate}

After the reorganization the whole dataset consisting of 2,337
motion sequences and 613,377 frames is divided into 65
actions\footnote{See the appendix for the
complete list
of 65 actions}. 

\subsection{Settings}
\noindent We used 10-fold cross validation to assess the performance
of a classifier. The data was randomly split into 10
balanced partitions of sequences. 
PO feature was formed by $5$ joints:head, hands and feet. 
The parameter $m$ in TD was set as  
$0.3$ second interval between frames. 
The total dimension of the feature vector is $33$. 
To test the distinctiveness of the features, we reported the 
classification accuracy for each frame, and evaluated the 
system performance by the accuracy of each sequence. 
The standard deviations were also calculated for 10-fold 
cross validation. 

We trained deep neural networks having two hidden layers of
sizes 1000 and 500 with rectified linear units\footnote{The
activation of a rectified linear unit is $\max(0, \alpha)$,
where $\alpha$ is the input to the unit.}. A learning rate
was selected automatically by a recently proposed ADADELTA
\cite{Zeiler2012}. 
Usually the optimal $\lambda$ can be selected by the
validation set and on a grid search. To illustrate the
influences of $\lambda$ in hybrid MLP, we selected four
different values for $\lambda$: $0$, $0.1$, $0.5$ and $0.9$.
The parameters were simply initialized randomly, and no
pretraining strategy was used\footnote{We used a publicly available
\textsc{Matlab} toolbox \mbox{\textit{deepmat}} for training and
evaluating deep
neural networks: https://github.com/kyunghyuncho/deepmat}.

When a tested classifier outputs the posterior probability
of a class given a frame, we chose the class of a sequence
by 
\[
\argmax_c \sum_{i=1}^N \log p(f_{i,c} = 1\mid \vf_i)
\]
based on Eq.~\eqref{eq:seq_prob_simple}. If a classifier
does not return a probability but only the chosen class, we
used a simple majority voting.

As a comparison, we tried an extreme learning machine (ELM)
\cite{Huang2006} and SVM in the same system. We used 2,000
hidden neurons for ELM. For SVM we used a radial-basis
function kernel, and the hyperparameters $C$ and
$\gamma$ were found through a grid-search and cross-validation.



\subsection{Qualitative Analysis: Visualization}

\noindent In order to have a better understanding of what a deep
neural network learns from the features, we visualized the
features using a deep autoencoder with two linear neurons in
the middle layer \cite{Hinton2006}. The deep autoencoder had
three hidden layers of size 1000, 500 and 100 between the
input and middle layers.  It should be noted that no label
information was used to train these deep autoencoders.  In
the experiment, we trained two deep autoencoders using with
and without the normalized trajectories (NT) to see what the
relative feature (PO+TD) provides to the system and the
impact of the absolute feature.
Since in previous works PCA has been often used for
dimensionality reduction of motion features, we also tried
to visualize features using the two leading principal
components.

In Fig.~\ref{fig:vis_rotate}, we visualized three distinct,
but very similar, actions; \textit{rotateArmsRBackward},
\textit{rotateArmsBothBackward} and
\textit{rotateArmsLBackward}. These actions in the figure
were clearly distinguishable when the deep autoencoder was
used.  However, \textit{rotateArmsRBackward} and
\textit{rotateArmsLBackward} were not distinguishable at all
when only the PO and TD features were used by PCA (see
Fig.~\ref{fig:vis_rotate} (c)). Even when all three features
(PO+TD+NT) were used, the visualization by PCA did not help
distinguishing these actions clearly.

In Fig.~\ref{fig:vis_jog}, two actions,
\textit{jogLeftCircle} and \textit{jogRightCircle}, were
visualized. 
When only PO and TD features were used, neither the deep
autoencoder nor PCA was able to capture differences
between those actions. However, the
deep autoencoder was able to distinguish those actions
clearly when all three proposed features were used (see
Fig.~\ref{fig:vis_jog} (b)). 

The former visualization shows that a deep neural network
with multiple nonlinear hidden layers can learn more
discriminative structure of data. 
Furthermore, according to the latter visualization, we can
see that the normalized trajectories help distinguish
locomotions with different traces, however, with only a
powerful model as a deep neural network.  Through the
experiment we could see that deep neural networks are able
to learn highly discriminative information from our features
of motion.

\subsection{Quantitative Analysis: Recognition}

\begin{table*}[t]
\caption{Classification accuracies. Standard deviations are
    shown inside brackets. 
    The highest accuracy in each row is marked bold.}
    \centering
    \begin{tabular}{c || c | c | c | c | c | c}
        Feature &     &     &     &
        \multicolumn{3}{c}{Hybrid MLP} \\
        Set & ELM & SVM & MLP & $\lambda=0.1$ & $0.5$ & $0.9$ \\
        \hline
        \hline
        \scriptsize PO+TD & 70.40\% {\scriptsize (1.32)} & 83.82\%
        {\scriptsize (0.79)} & 84.35\% {\scriptsize (0.91)} &
        84.39\% {\scriptsize (0.87)} & \textbf{84.57}\% {\scriptsize
        (1.56)} & 84.23\% {\scriptsize (1.27)} \\
        \hline
        \scriptsize PO+TD+NT & 74.28\% {\scriptsize (1.56)} & 87.06\%
        {\scriptsize (0.82)} & 87.42\% {\scriptsize (1.43)} &
        \textbf{87.96}\% {\scriptsize (1.38)} & 87.34\% {\scriptsize
        (0.66)} & 87.28\% {\scriptsize (1.38)} \\
    \end{tabular}
    \vspace{2mm}
    \label{tbl:frame_acc}
    \centering
    \begin{tabular}{c || c | c | c | c | c | c}
        Feature &     &     &     &
        \multicolumn{3}{c}{Hybrid MLP} \\
        Set & ELM & SVM & MLP & $\lambda=0.1$ & $0.5$ & $0.9$ \\
        \hline
        \hline
        \scriptsize PO+TD & 91.57\% {\scriptsize (0.88)} & 94.95\%
        {\scriptsize (0.82)} & 95.20\% {\scriptsize (1.38)} &
        95.46\% {\scriptsize (0.99)} & \textbf{95.59}\% {\scriptsize
        (0.76)} & 95.55\% {\scriptsize (1.14)} \\
        \hline
        \scriptsize PO+TD+NT & 92.76\% {\scriptsize (1.53)} & 95.12\%
        {\scriptsize (0.58)} & 94.86\% {\scriptsize (0.99)} &
        \textbf{95.21}\% {\scriptsize (0.86)} & 94.82\% {\scriptsize
        (1.17)} & 95.04\% {\scriptsize (0.86)} \\
    \end{tabular}
    \vspace{2mm}
    \label{tbl:seq_acc}
\end{table*}
\noindent In Tab.~\ref{tbl:frame_acc}, the frame-level accuracies
obtained by various classifiers with two different sets of
features can be found. We can see that 
the NT feature clearly increases the classification accuracy
around $3-4$\% for all the classifiers.  Comparing the
different classifiers, we can see that the MLPs were able to
obtain significantly higher accuracies than the ELM and
perform slightly better than SVM. Furthermore, although it
is not clearly significant statistically, we can see that a
hybrid MLP often outperforms the regular MLP with a right
choice of $\lambda$. 

A similar trend of the MLPs outperforming the other
classifiers could be observed also in sequence-level
performance shown in Tab.~\ref{tbl:seq_acc}.  Again in the
sequence-level classification, we observed that the hybrid
MLP with a right choice of $\lambda$ marginally outperformed
the regular MLP, and it also outperformed SVM and ELM. 
For both frame-level and sequence-level accuracy, the
highest accuracy for PO+TD features is from hybrid MLP with
$\lambda=0.5$ and for the whole feature set with
$\lambda=0.1$. 


However, the classification accuracies obtained using 
the two sets of features (PO+TD vs PO+TD+NT) are very close
to each other.
Compared to the $3-4$\% differences in the classification
for each frame, the differences between the performance
obtained using the two sets are within the standard
deviations. 
Even though NT feature increases the frame accuracy significantly 
it did not have the same effect on the sequence level. 
One potential reason is that 
once a certain level of frame level recognition is achieved, 
the sequence level performance using our posterior
probability model saturates. 


\section{\uppercase{Conclusions}}

\noindent In this paper, we proposed a gesture recognition system
using multi-layer perceptrons for recognizing motion
sequences with novel features based on relative joint
positions (PO), temporal differences (TD) and normalized
trajectories (NT). 

The experiments with a large motion capture dataset (HDM05)
revealed that (hybrid) multi-layer perceptrons could achieve
higher recognition rate than there is the other classifiers could, for 65 
classes with an accuracy of above 95\%. 
Furthermore, the visualization of feature set of the motion sequences
by deep autoencoders showed the effectiveness of the
proposed feature sets and enabled us to study what deep
neural networks learned.
Interestingly, a powerful model like a deep neural network
combined with an informative feature set was able to capture
the discriminative structure of motion sequences, which was
confirmed by both the recognition and visualization
experiments.  This suggests that a deep neural network is
able to extract highly discriminative features from motion
data.

One limitation of our approach is that temporal independence
was assumed when combining the per-frame posterior
probabilities in a sequence. In future it will be
interesting to investigate possibilities of modeling
temporal dependence.

\section*{Acknowledgments}

This work was funded by Aalto MIDE programme (project
UI-ART), Multimodally grounded language technology (254104)
and Finnish Center of Excellence in Computational Inference
Research COIN (251170) of the Academy of Finland.

\bibliographystyle{apalike}
\bibliography{nips,motion}

\newpage
\section*{\uppercase{Appendix}: 65 Actions in HDM05 Dataset}
{
\small 
\begin{enumerate}
\item cartwheelLHandStart1Reps\\cartwheelLHandStart2Reps\\cartwheelRHandStart1Reps
\item clap1Reps \\clap5Reps 
\item clapAboveHead1Reps \\clapAboveHead5Reps 
\item depositFloorR 
\item depositHighR 
\item depositLowR 
\item depositMiddleR 
\item elbowToKnee1RepsLelbowStart\\elbowToKnee1RepsRelbowStart\\elbowToKnee3RepsLelbowStart\\elbowToKnee3RepsRelbowStart
\item grabFloorR 
\item grabHighR 
\item grabLowR 
\item grabMiddleR 
\item hitRHandHead 
\item hopBothLegs1hops\\hopBothLegs2hops\\hopBothLegs3hops 
\item hopLLeg1hops\\hopLLeg2hops\\hopLLeg3hops 
\item hopRLeg1hops\\hopRLeg2hops\\hopRLeg3hops 
\item jogLeftCircle4StepsRstart\\jogLeftCircle6StepsRstart 
\item jogOnPlaceStartAir2StepsLStart\\jogOnPlaceStartAir2StepsRStart\\jogOnPlaceStartAir4StepsLStart\\jogOnPlaceStartFloor2StepsRStart\\jogOnPlaceStartFloor4StepsRStart 
\item jogRightCircle4StepsLstart\\jogRightCircle4StepsRstart\\jogRightCircle6StepsLstart\\jogRightCircle6StepsRstart
\item jumpDown 
\item jumpingJack1Reps\\jumpingJack3Reps 
\item kickLFront1Reps\\kickLFront2Reps 
\item kickLSide1Reps\\kickLSide2Reps 
\item kickRFront1Reps\\kickRFront2Reps 
\item kickRSide1Reps\\kickRSide2Reps 
\item lieDownFloor 
\item punchLFront1Reps\\punchLFront2Reps 
\item punchLSide1Reps\\punchLSide2Reps 
\item punchRFront1Reps\\punchRFront2Reps 
\item punchRSide1Reps\\punchRSide2Reps 
\item rotateArmsBothBackward1Reps\\rotateArmsBothBackward3Reps 
\item rotateArmsBothForward1Reps\\rotateArmsBothForward3Reps 
\item rotateArmsLBackward1Reps\\rotateArmsLBackward3Reps 
\item rotateArmsLForward1Reps\\rotateArmsLForward3Reps 
\item rotateArmsRBackward1Reps\\rotateArmsRBackward3Reps 
\item rotateArmsRForward1Reps\\rotateArmsRForward3Reps 
\item runOnPlaceStartAir2StepsLStart\\runOnPlaceStartAir2StepsRStart\\runOnPlaceStartAir4StepsLStart\\runOnPlaceStartFloor2StepsRStart\\runOnPlaceStartFloor4StepsRStart
\item shuffle2StepsLStart\\shuffle2StepsRStart\\shuffle4StepsLStart\\shuffle4StepsRStart
\item sitDownChair 
\item sitDownFloor 
\item sitDownKneelTieShoes 
\item sitDownTable 
\item skier1RepsLstart\\skier3RepsLstart 
\item sneak2StepsLStart\\sneak2StepsRStart\\sneak4StepsLStart\\sneak4StepsRStart
\item squat1Reps\\squat3Reps 
\item staircaseDown3Rstart 
\item staircaseUp3Rstart 
\item standUpKneelToStand 
\item standUpLieFloor 
\item standUpSitChair 
\item standUpSitFloor 
\item standUpSitTable 
\item throwBasketball 
\item throwFarR 
\item throwSittingHighR\\throwSittingLowR 
\item throwStandingHighR\\throwStandingLowR 
\item turnLeft 
\item turnRight 
\item walk2StepsLstart\\walk2StepsRstart\\walk4StepsLstart\\walk4StepsRstart 
\item walkBackwards2StepsRstart\\walkBackwards4StepsRstart 
\item walkLeft2Steps\\walkLeft3Steps 
\item walkLeftCircle4StepsLstart\\walkLeftCircle4StepsRstart\\walkLeftCircle6StepsLstart\\walkLeftCircle6StepsRstart
\item walkOnPlace2StepsLStart\\walkOnPlace2StepsRStart\\walkOnPlace4StepsLStart\\walkOnPlace4StepsRStart
\item walkRightCircle4StepsLstart\\walkRightCircle4StepsRstart\\walkRightCircle6StepsLstart\\walkRightCircle6StepsRstart
\item walkRightCrossFront2Steps\\walkRightCrossFront3Steps 
\end{enumerate}
}


\vfill
\end{document}